\documentclass{article}

\usepackage{PRIMEarxiv}
\usepackage{framed,multirow}
\usepackage{amssymb}
\usepackage{latexsym}
\usepackage{amsmath}
\usepackage{subfigure}
\usepackage{multirow}
\usepackage{threeparttable}
\usepackage{hyperref}

\usepackage[utf8]{inputenc} 
\usepackage[T1]{fontenc}    
\usepackage{hyperref}       
\usepackage{url}            
\usepackage{booktabs}       
\usepackage{amsfonts}       
\usepackage{nicefrac}       
\usepackage{microtype}      
\usepackage{lipsum}
\usepackage{fancyhdr}       
\usepackage{graphicx}       
\graphicspath{{media/}}     

\title{An Efficient Framework for Few-shot Skeleton-based Temporal Action Segmentation}

\author{Leiyang Xu\\
	Department of Control Science and Engineering\\
	Harbin Institute of Technology\\
	Harbin, China\\
	\texttt{xuleiyang@stu.hit.edu.cn} \\
	\And
	Qiang Wang \\
	Department of Control Science and Engineering\\
	Harbin Institute of Technology\\
	Harbin, China\\
	\texttt{wangqiang@hit.edu.cn} \\
	\And
	Xiaotian Lin \\
	Department of Control Science and Engineering\\
	Harbin Institute of Technology\\
	Harbin, China\\
	\texttt{xiaotian.lin@hit.edu.cn} \\
	\And
	Lin Yuan\\
	Department of Control Science and Engineering\\
	Harbin Institute of Technology\\
	Harbin, China\\
	\texttt{\_eunseo\_v@hit.edu.cn} \\\\
}

\begin{document}
\maketitle

\begin{abstract}
Temporal action segmentation (TAS) aims to classify and locate actions in the long untrimmed action sequence. With the success of deep learning, many deep models for action segmentation have emerged. However, few-shot TAS is still a challenging problem. This study proposes an efficient framework for the few-shot skeleton-based TAS, including a data augmentation method and an improved model. The data augmentation approach based on motion interpolation is presented here to solve the problem of insufficient data, and can increase the number of samples significantly by synthesizing action sequences. Besides, we concatenate a Connectionist Temporal Classification (CTC) layer with a network designed for skeleton-based TAS to obtain an optimized model. Leveraging CTC can enhance the temporal alignment between prediction and ground truth and further improve the segment-wise metrics of segmentation results. Extensive experiments on both public and self-constructed datasets, including two small-scale datasets and one large-scale dataset, show the effectiveness of two proposed methods in improving the performance of the few-shot skeleton-based TAS task.
\end{abstract}


\section{Introduction}\label{sec1}
Human activity analysis is an important research field in computer vision. Its applications include surveillance systems \cite{dave2022gabriellav2,elharrouss2021combined}, medical rehabilitation \cite{yin2021mc,wang2021temporal,hossain2021human}, and physical training \cite{chen2007physics,hao2021recognition}. While methods for classifying short trimmed actions have achieved high accuracy \cite{shahroudy2016ntu,ren2020survey, plizzari2021skeleton,cao2021few}, it is still challenging to both identify and localize events in long untrimmed motion sequence.

With the development and availability of low-cost and high-accuracy motion capture systems, it is convenient to collect skeleton-based motion sequences. Motion capture sequences represent human motions by recording the time series of human joint movement and rotation. This representation has the advantage of robustness to background and illumination. For modeling 3D skeletal data, one of the most important assignments is to extract suitable spatial and temporal features. In recent years, there have many excellent works on skeleton-based action recognition. Most of these action recognition tasks used the trimmed actions, which have been segmented and labeled, as research subjects. However, human activity usually appears as untrimmed motion sequences instead of single actions.
\begin{figure*}[!t]
  \centering
  \includegraphics[width=1.0\textwidth]{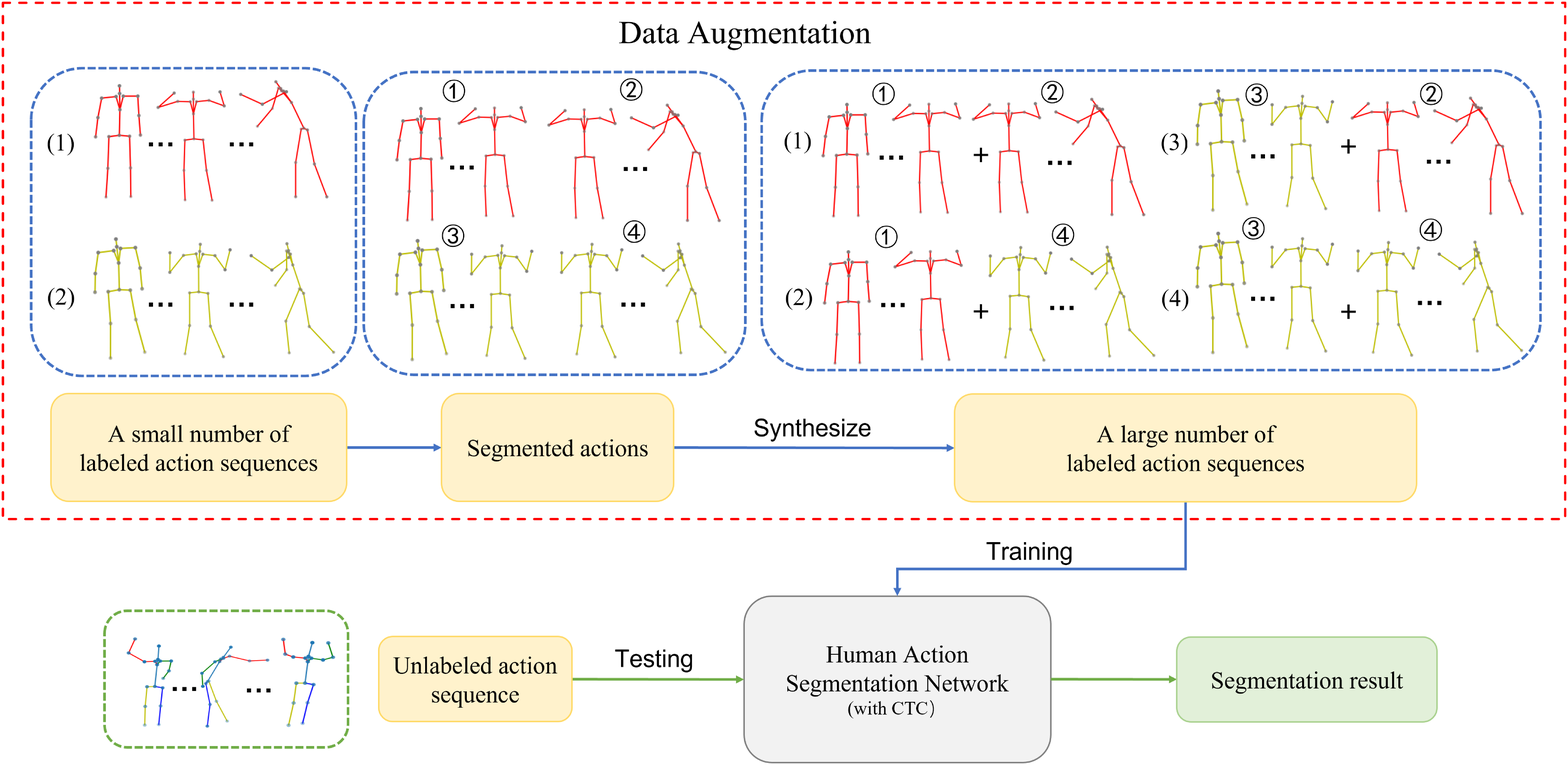}
  \caption{Overview of our proposed framework for few-shot skeleton-based temporal action segmentation.}\label{fig1}
\end{figure*}

Temporal Action Segmentation (TAS) aims to recognize actions in the motion sequence and localize the start frame and end frame of each action. Prior works \cite{carrara2019lstm,xu2021long,pang2019deep} prefer to exploit various networks suitable for time-series, such as RNN, and LSTM, to model temporal context. For skeletal motion data, it is also necessary to extract the spatial features of the skeleton. \cite{song2018spatio} propose a spatial and temporal attention model to explore the spatial and temporal discriminative features. Their networks are based on recurrent neural networks with long short-term memory units. \cite{wang2018beyond} introduce three primitives geometries (joints, edges, and surfaces) to leverage the geometric relations and design an end-to-end RNN-based network to perform frame-wise action classification and detection. Many methods based on deep convolutional neural networks, including the above methods, have achieved good results on action segmentation tasks. However, these networks are heavily reliant on big data to avoid overfitting. Unfortunately, labeling large amounts of skeletal motion data is expensive and time-consuming. Sometimes, it is also difficult to collect sufficient training data for training a deep learning model, such as motions in some professional fields. Data augmentation is one of the commonly used methods to address this problem.

Data augmentation is a technique used to increase the amount of data by adding slightly modified copies of already existing data or newly created synthetic data from existing data \cite{li2022data}, which is a data-space solution to insufficient data size such that various deep learning models can be built. These methods have been successfully used in many fields, such as image analysis tasks \cite{shorten2019survey,wang2017effectiveness,yoo2020rethinking}, video processing \cite{yun2020videomix,cauli2022survey,zhang2020self} and action recognition \cite{huynh2019encoding,xu2016multi,meng2019sample}. Nevertheless, the data augmentation method specially designed for skeleton-based action sequences is rare. \cite{yao2022multi} propose a strategy to augment the data with different speeds and skeleton sizes. However, this method does not add new action sequences but fine-tunes the existing action sequences. In this work, we present a data augmentation method for few-shot skeleton-based temporal action segmentation (FS-TAS) task, which augments labeled data by synthesizing new action sequences.

In this paper, we propose two approaches to improve FS-TAS results: data augmentation and adding Connectionist Temporal Classification (CTC) in the baseline architecture. First, to solve the problem of limited data, we synthesize new labeled action sequences based on a small amount of labeled data and merge synthetic action sequences with existing data into a new synthetic dataset. Second, the synthetic dataset is used as training data to train the skeleton-based TAS (S-TAS) network. Here, we quote Multi-Stage Spatial-Temporal Graph Convolutional Networks (MS-GCN) proposed in \cite{filtjens2022skeleton} as a simple baseline architecture. Based on MS-GCN, we add CTC to improve the temporal alignment between network predictions and ground truth. Finally, we experiment on three datasets, including two synthetic datasets and one large-scale public dataset. Two synthetic datasets are used to verify the effectiveness of our proposed data augmentation method, and all three datasets are used to prove that the addition of CTC can improve the performance of MS-GCN. An overview of our approach is sketched in Figure 1.
The main contributions of our work can be summarized as follows:

(1) A data augmentation method is proposed to augment training data by synthesizing new action sequences. In this way, deep models can be applied in FS-TAS tasks.

(2) CTC is cascaded together with a baseline network to form a combined network that can be trained end-to-end and outperform on FS-TAS tasks.

(3) The proposed framework has been evaluated on three datasets, including two small-scale datasets and one large-scale public dataset. Experiment results show the above two methods are effective.

\section{Related Works}
\label{sec2}
In this section, we review recent studies related to our method, which can be divided into two parts: TAS and CTC.

\subsection{Temporal Action Segmentation (TAS)}
TAS requires not only recognizing the actions in the sequence but also localizing their start and end frames. Compared with action recognition, TAS is closer to real-world application scenarios. According to the form of action data, TAS is mainly divided into two types: video-based \cite{jiang2014thumos,kuehne2014language,bojanowski2014weakly,ng2021weakly} and skeleton-based \cite{sung2012unstructured,liu2017pku,yun2012two}. \cite{farha2019ms} proposed a Multi-Stage Temporal Convolutional Network (MS-TCN) for video-based action segmentation. Each stage generates an initial prediction that is refined by the next stage. This model has achieved state-of-the-art on some challenging datasets. In contrast with video, skeletal action data has higher dimensions and retains more information, including spatial features between joints. \cite{yan2018spatial} proposed a model for skeleton-based action recognition called Spatial-Temporal Graph Convolutional Networks (ST-GCN), which can automatically learn both spatial and temporal patterns from data. The presented model combines GCN and TCN for the first time and works well on many datasets. Based on the above two excellent works, \cite{filtjens2022skeleton} introduced MS-GCN. The idea of this framework is to amalgamate the best practices in the convolutional network for video-based TAS and skeleton-based action recognition design into a simple baseline architecture for S-TAS. Experiments show that the performance of MS-GCN on S-TAS datasets is better than ST-GCN and MS-TCN. The above methods have achieved excellent performance on TAD, but this relies on a large amount of training data. An insufficient amount of data can lead to severe overfitting of the model. In this work, we propose a data augmentation method for action sequences, which make it possible to apply deep models on few-shot S-TAD tasks.
\subsection{Connectionist Temporal Classification (CTC)}
CTC \cite{graves2006connectionist} is a way to get around not knowing the alignment between the input and the output \cite{hannun2017sequence}, which is widely used in many applications like speech recognition \cite{kim2017joint,zhang2019investigation} and machine translation \cite{libovicky2018end}. Similarly, gesture or action segmentation in temporal sequences is another suitable application area for CTC. \cite{molchanov2016online} proposed a network based on recurrent 3D convolutional neural networks for online gestures detection. They leveraged CTC to train the network to predict class labels, enabling gesture classification based on the core stages of gestures without explicit pre-segmentation. \cite{li2020key} also employed CTC and LSTM for sign language recognition to balance the sequence alignment and dependency. \cite{huang2016connectionist} adopt CTC for action labeling in video. They supervised CTC training by prior knowledge and explicitly enforced consistency with frame-to-frame visual similarities. These works verify that CTC can solve the problem of time series alignment to a certain extent. Therefore, we apply CTC to the S-TAS task similar to the above tasks and enhance the temporal alignment of MS-GCN by adding CTC to the network structure.

\section{Data Augmentation}\label{sec3}

As mentioned before, data augmentation aims at increasing the amount of training data to avoid the overfitting of deep learning models. For the FS-TAS task, data augmentation task is to generate new labeled action sequences from existing data.

\subsection{Motion Interpolation}
Motion interpolation applies in many situations, such as motion smoothing and keyframe animation. In the process of synthesizing action sequences, we also need the motion interpolation method to realize the splicing between different actions. For the human skeleton, spherical linear interpolation (slerp) \cite{shoemake1985animating} is commonly used for motion interpolation. Slerp is a quaternion linear interpolation used to smooth the difference between two quaternions representing rotations. Support two points are represented as two quaternions \(q_{1}\) and \(q_{2}\), slerp defines as:
\begin{equation}
Slerp(q_{1},q_{2};t) = \frac{sin(1-t)\theta}{sin\theta}q_{1} + \frac{sint\theta}{sin\theta}q_{2}
\end{equation}
where the dot produce \(q_{1}\cdot q_{2}=cos\theta\). The parameter \(t\) moves from 0 (the expression equals to \(q_{1}\)) to 1 (the expression equals to \(q_{2}\)).

\begin{figure*}[!t]
  \centering
  \includegraphics[width=0.8\textwidth]{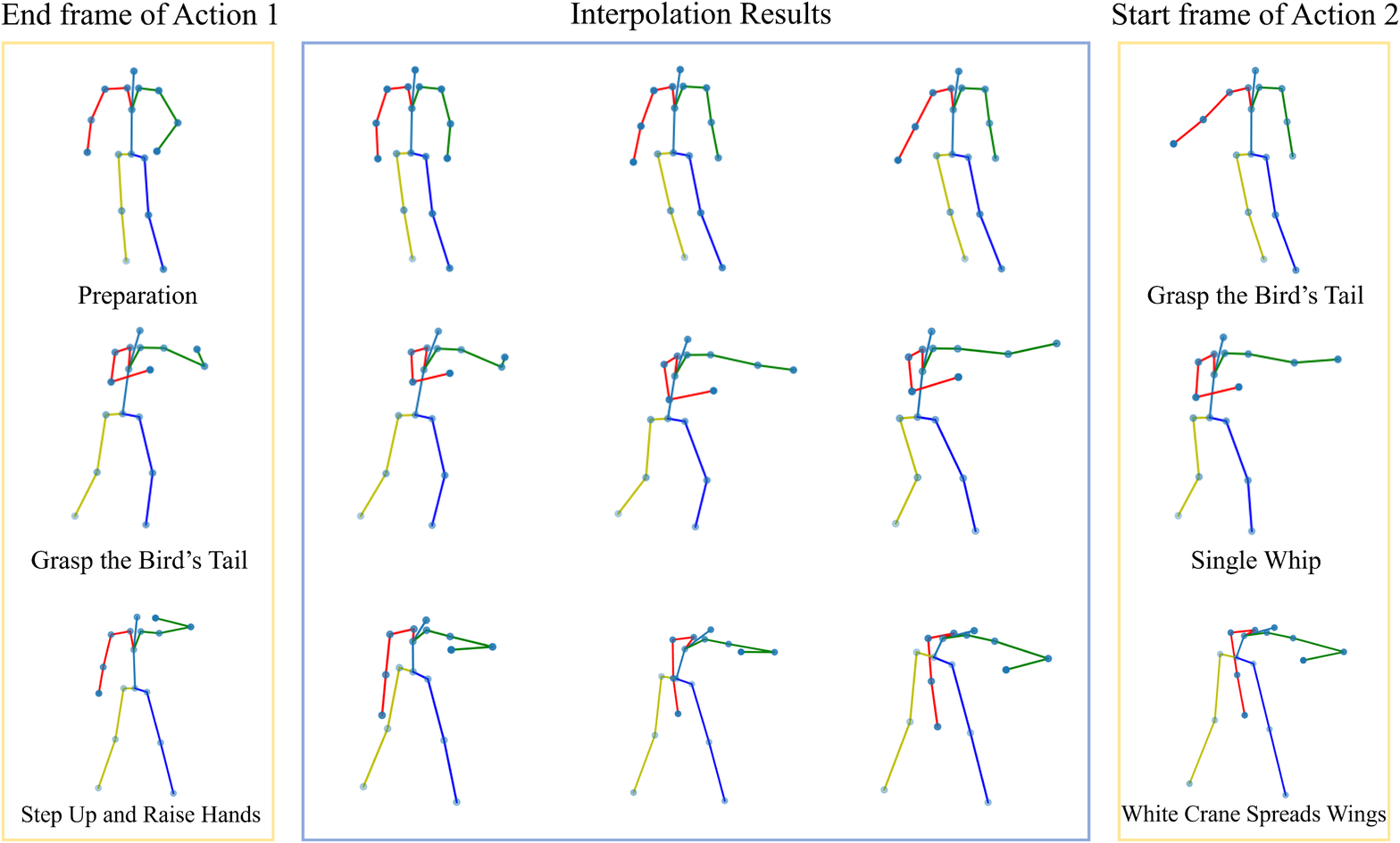}
  \caption{Interpolation results between two actions on three different action sequences.}
  \label{fig2}
\end{figure*}

When performing slerp for motion interpolation, we view joints as independent variants, which may not consider the relationship between joints. However, given the small amount of data and the high similarity between the two human poses being interpolated, slerp can perform well in our task. Figure 2 illustrates the interpolation results obtained by slerp. Two yellow boxes are the end frame of action 1 and the start frame of action 2, and the blue boxes are the interpolated human poses.
\subsection{Synthetic Action Sequences}

Suppose some action sequences that contain fixed category action primitives. They can be performed by different subjects or by the same subject multiple times. Then, we introduce how to synthesize large amounts of new action sequences. Figure 3 shows the flow of synthesizing the new action sequences.

\begin{figure*}[!t]
  \centering
  \includegraphics[width=1.0\textwidth]{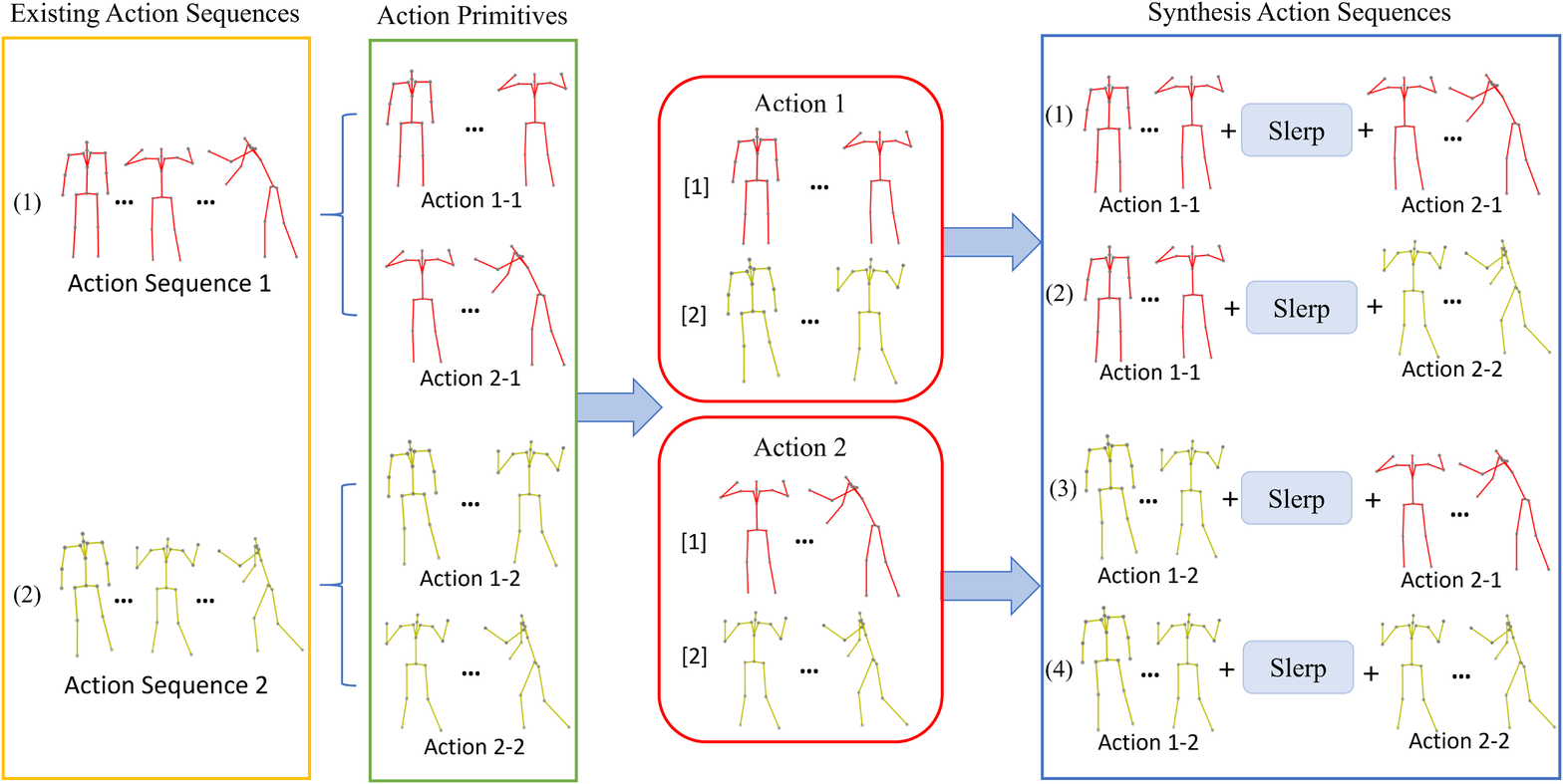}
  \caption{The flow of synthesizing the new action sequences.}
  \label{fig3}
\end{figure*}

First, divide the existing action sequences into multiple action primitives by labels. Second, separate the action primitives by category. Third, we select one action primitive in each category and use slerp to concatenate action primitives in the order they appear in the action sequence. In particular, interpolation is also necessary even if two connected action primitives are from the same action sequence to ensure the continuity of coordinates. In this way, training data can increase at an exponential rate by synthesizing new action sequences. Assume there are \(M\) labeled action sequences, and each action sequence has \(n\) action primitives. Theoretically, the number of synthetic action sequences is:
\begin{equation}
Num = M^{n}
\end{equation}

In our experiment, it is sufficient to use only two action sequences (\(M=2\)) as training data. Each action sequence contains ten action primitives (\(n=10\)). After data augmentation, training data augment to 1024 (\(2^{10}\)) samples. If the start pose and the end pose of each action primitive is similar, the order of the action primitives may not be considered when synthesizing the action sequence. We will analyze this possibility and the label of interpolated frames in the experiment section.

\section{End-to-End Action Segmentation}
\label{sec4}
In this section, we introduce the basic network architecture proposed in \cite{filtjens2022skeleton} and describe the improvement we have made on this basis.
\subsection{Network Architecture}
MS-GCN was selected as the baseline network for skeleton-based action segmentation. It combines the best practices from MS-TCN \cite{farha2019ms}, and ST-GCN \cite{yan2018spatial}. The network architectures of these basic models are visualized in Figure 4.

\begin{figure*}[!t]
  \centering
  \includegraphics[width=1.0\textwidth]{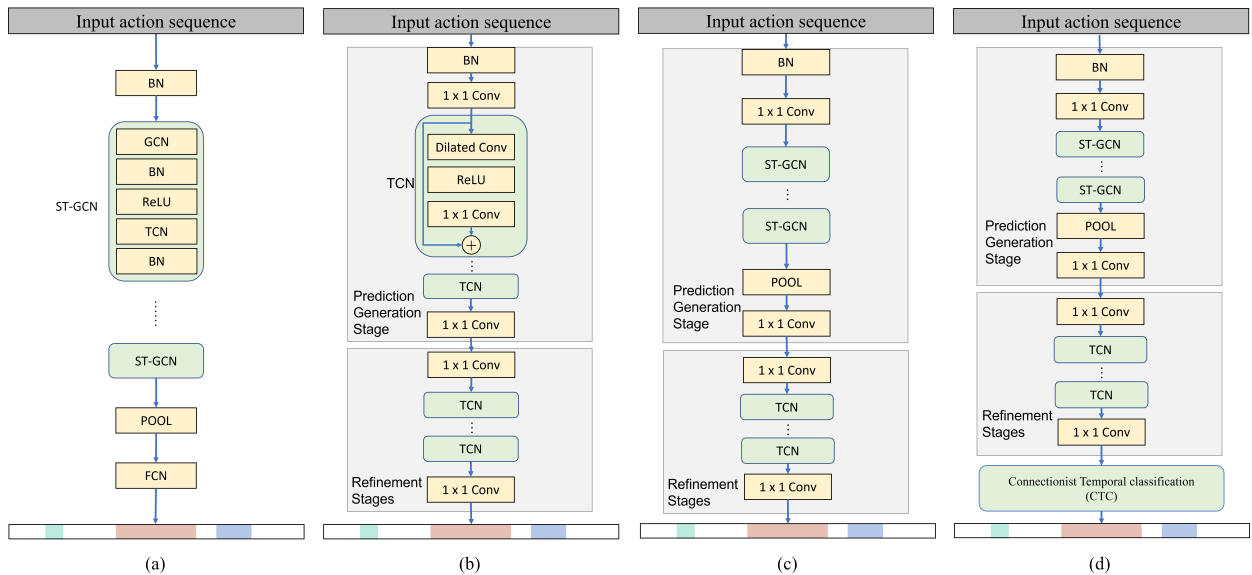}
  \caption{Architecture of our model and three baseline models. (a) ST-GCN. (b) MS-TCN. (c) MS-GCN. (d) Our model.}
  \label{fig:4}
\end{figure*}

In the prediction generation stage, we begin to extract the features of skeleton by ST-GCN, which treats human skeleton as a kind of graph and learns a representation on this graph. Each ST-GCN block can be described as:
\begin{equation}
f_{out}=ReLU(BN(W_{1}\ast(\sum_{p}W_{p}(f_{in}A_{p})\otimes M_{p})+b_{1}))
\end{equation}
where \(\ast\) denotes the convolution operator, \(\otimes\) denotes the dot product operator; \(f_{in}\in\mathbb{R}^{C_{in}\times T\times N}\) is the input of the block, \(T\) are the number of frames, \(N\) are the number of joints, \(C_{in}\) is the number of input feature channels; \(A_{p}\) is the \(N\times N\) adjacency matrix, which represents the connection relationship between joints; \(W_{p}\in\mathbb{R}^{C_{out}\times C_{in}\times 1\times 1}\) is a weight matrix; \(M_{p}\) is the \(N\times N\) learnable attention mask, which represents the importance of each joint; \(W_{1}\) is a \(k\times 1\times C_{out}\times C_{out}\) weight matrix with kernel size \(k\), \(b_{1}\in\mathbb{R}^{C_{out}}\), and \(f_{out}\in\mathbb{R}^{C_{out}\times T\times N}\).

In the refinement stages, we perform several TCN blocks to refine the frame-wise features output from the prediction stage. Multi-stage temporal convolutional neural network (MS-TCN) can reduce the over-segmentation error. Each TCN block is formalized as:
\begin{equation}
f_{out}=f_{in}+W_{2}*(ReLU(W_{1}*f_{in}+b_{1}))+b_{2}
\end{equation}
where \(\ast\) denotes the convolution operator; \(f_{in}\in\mathbb{R}^{T\times  C}\) is the input of this block or the output of the previous block, \(T\) are the number of frame, \(C\) is the number of input feature channels; \(W_{1}\in\mathbb{R}^{C\times C\times k}\) are the weights of the dilated convolution filters with kernel size k; \(W_{2}\in\mathbb{R}^{C\times C\times 1}\) are the weights of a \(1\times 1\) convolution, \(b_{1},b_{2}\in\mathbb{R}^{C}\), and \(f_{out}\in\mathbb{R}^{T\times C}\).

For MS-TCN, each stage takes an initial prediction from the previous stage and refines it. The process is defined as:
\begin{equation}
Y^{s} = \mathcal{F}(Y^{s-1})
\end{equation}
where \(Y^{s}\) is the output at block \(s\), \(Y^{s-1}\) is the output at block \(s-1\), and \(\mathcal{F}\) is a TCN block defined in Equation (4).

ST-GCN is an excellent method for skeleton-based action recognition, and MS-TCN overperforms on video-based action segmentation. MS-GCN combines the advantages of the above two models. It employs the ST-GCN block in the prediction stage, which can learn both spatial and temporal patterns from skeleton data and maintains the structure of MS-TCN in the refinement stages, which helps alleviate over-segmentation error. This allows MS-GCN to perform well on skeleton-based action segmentation tasks. Therefore, we select MS-GCN as a baseline network to validate our proposed data augmentation method.

\subsection{Loss Function}
MS-GCN and MS-TCN use the same loss function, which combines cross-entropy loss and mean squared error loss.

Cross-entropy (CE) loss measures the accuracy loss of classification:
\begin{equation}
\mathcal{L}_{cls}=\frac{1}{T}\sum_{t}-y_{t,l}\log(\hat y_{t,l})
\end{equation}
where \(y_{t,l}\) and \(\hat y_{t,l}\) are the ground truth label and the predicted probability for label \(l\) at time \(t\).

Mean squared error (MSE) loss guarantees the smoothing loss of segmentation results:
\begin{equation}
\mathcal{L}_{T-MSE}=\frac{1}{TL}\sum_{t,l}\tilde \Delta^{2}_{t,l}
\end{equation}
\begin{equation}
\tilde \Delta_{t,l}= 
\begin{cases}
\Delta_{t,l} & :\Delta_{t,l}\leq\tau \\
\tau & :otherwise \\
\end{cases}
\end{equation}
\begin{equation}
\Delta_{t,l}=\left\vert \log(\hat y_{t,l})-\log(\hat y_{t-1,l}) \right\vert
\end{equation}
where \(T\) is the length of action sequence, \(L\) is the number of classes, and \(\hat y_{t,l}\) is the probability of class \(l\) at time \(t\).

The loss function for MS-GCN and MS-TCN is a combination of CE loss and MSE loss:
\begin{equation}
\mathcal{L}_{s}=\mathcal{L}_{cls}+\lambda\mathcal{L}_{T-MSE}
\end{equation}
where \(\lambda\) is a parameter to control the contribution of two losses.

CE loss strives to keep the prediction the same as the ground truth for each frame, ensuring that each frame can be classified correctly. MSE loss tried to make adjacent frames the same category to avoid over-segmentation as much as possible. However, the above two losses do not consider the alignment of the output and the ground truth at the sequence level. Therefore, we add CTC loss to the loss function. CTC loss relaxes the requirement for classification accuracy of transition frames between two action primitives, emphasizing the overall alignment of action sequences:
\begin{equation}
\mathcal{L}_{CTC}=-\ln\sum^{\left\vert z^{'} \right\vert}_{u=1}\alpha(t,u)\beta(t,u)
\end{equation}
\begin{equation}
\alpha(t,u)\beta(t,u)=\sum_{\pi\in X(t,u)}\prod^{T}_{t=1}y^{t}_{\pi_{t}}
\end{equation}
where \(y^{t}_{k}\) is the probability of output \(k\) at time \(t\), \(\pi\) is the output path, \(z\) is the label sequence, \(z^{'}\) is a modified label sequence with blanks added to the beginning and the end of \(z\).
\(A\) is a set containing all labels, \(A^{'}=A\cup\left\{ blank \right\}\), \(A^{'T}\) denotes the set of length \(T\) sequences over \(A\), function \(\mathcal{F}\) represents the mapping between paths and label sequences. More details can be found in \cite{graves2012supervised}.

The final loss function for our model (Figure 4(d)) is described as:
\begin{equation}
\mathcal{L}_{s}=\mathcal{L}_{cls}+\lambda\mathcal{L}_{T-MSE}+\mu\mathcal{L}_{CTC}
\end{equation}
where \(\lambda\) and \(\mu\) are two parameters to determine the contribution of CE loss, MSE loss, and CTC loss.

\subsection{Implementation Details}
Some model parameters follow the settings in MS-TCN. The model contains one prediction generation stage and three refinement stages, and each stage has ten layers. Each layer has 64 filters with a filter size of 3. The dilation factor is doubled at each layer. For the loss function, we add the CTC loss, so count a new parameter \(\mu\).  We set \(\tau = 4\), \(\lambda = 0.15\), and \(\mu = 0.0005\). For the optimizer, the Adam optimizer is utilized with a learning rate of 0.0005. In the experiments, we train the model with 100 epochs and set batch size as 4 for the Tai Chi dataset, and 8 for UTD-MHAD2 and PKU-MMD.

\section{Experiments}\label{sec5}
In this section, we design a series of experiments to identify our framework. We first perform ablation experiments on the proposed framework to confirm that the CTC layer and the data augmentation are adequate for the FS-TAS task. Then, we study the two proposed methods separately, including parameter settings and various comparative experiments.

\subsection{Datasets}
In the experiments, we use three datasets: two synthetic datasets generated by augmenting small-scale datasets and one large dataset. Two synthetic datasets are used to test the data augmentation method, and all three datasets are used to validate the effectiveness of CTC.

\begin{figure}[htbp]
	\centering
	\subfigure[Tai Chi]{
		\begin{minipage}[b]{0.2\textwidth}
    \centering
		\includegraphics[width=0.6\textwidth]{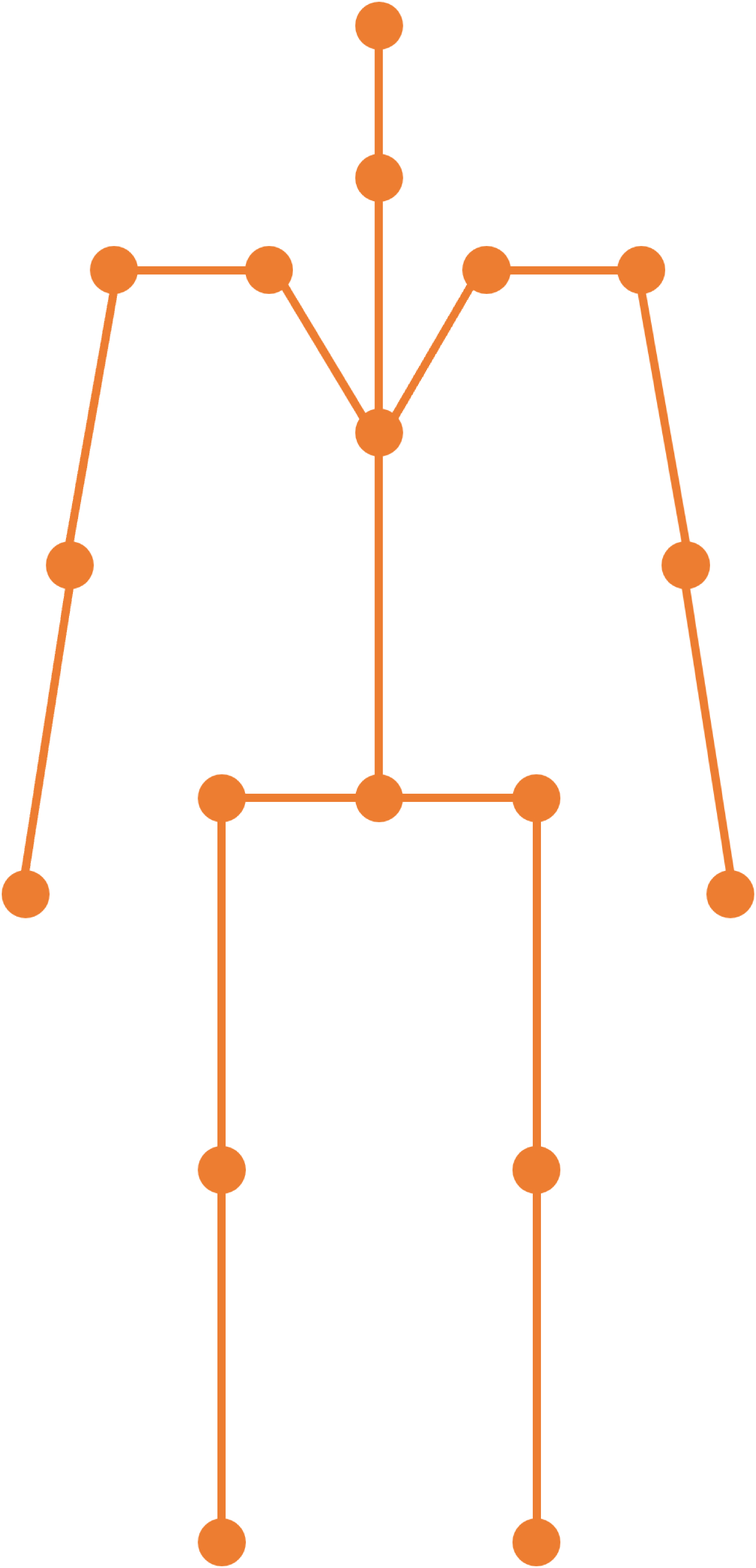}
		\end{minipage}
		\label{fig:5_1}
	}
    \subfigure[UTD-MHAD2 and PKU-MMD]{
    	\begin{minipage}[b]{0.3\textwidth}
            \centering
   		    \includegraphics[width=0.4\textwidth]{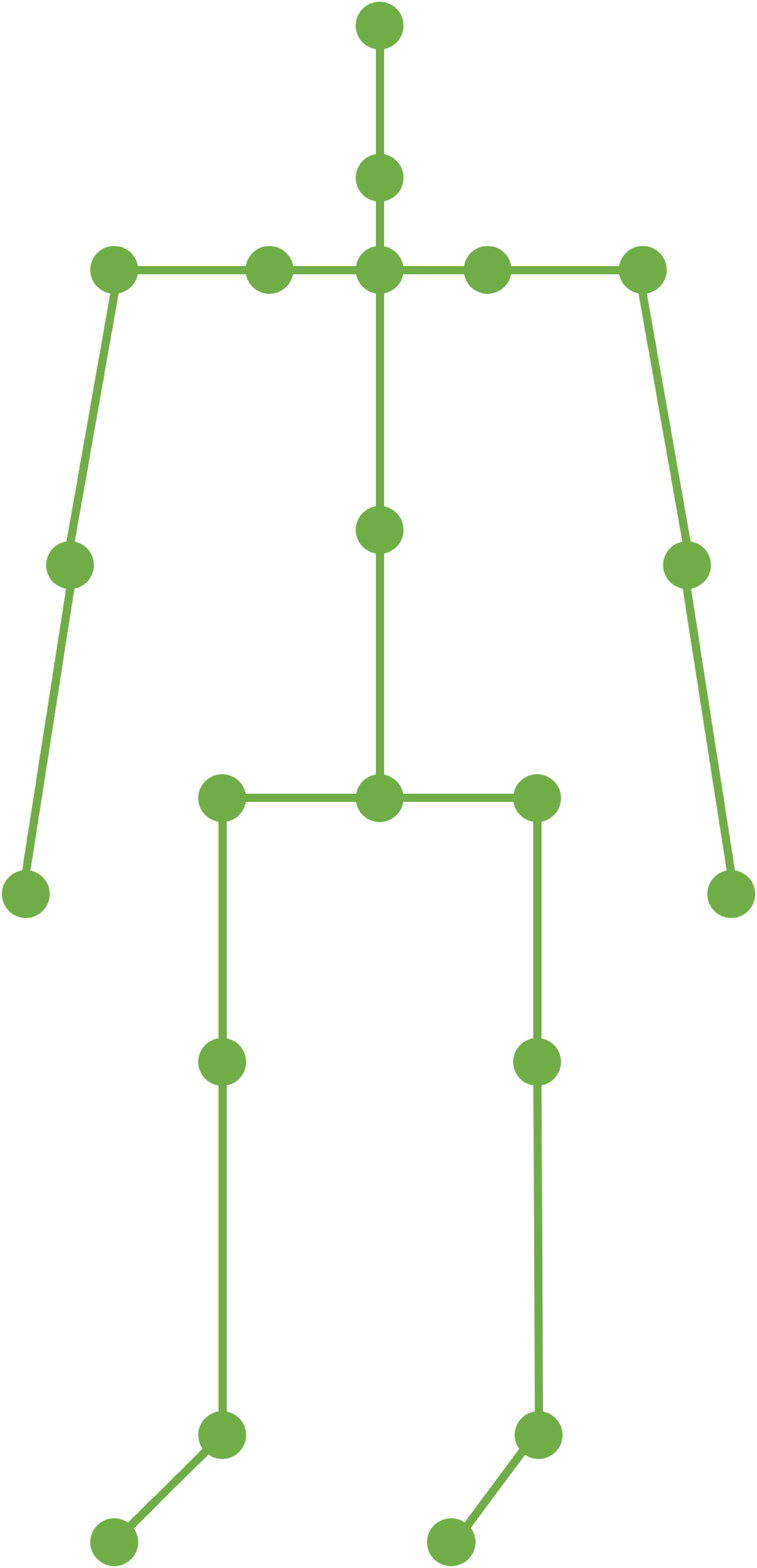}
    	\end{minipage}
	\label{fig:5_2}
    }
	\caption{Illustrations of human skeletons.}
	\label{fig:5}
\end{figure}

Considering small-scale datasets that require data augmentation, we divide them into the following two cases: (1) The dataset only contains a small amount of labeled action sequence data, which is insufficient for training data required by the deep model. Tai Chi dataset \cite{xu2020using} fits this case. (2) In the dataset, the data is stored as single action primitives instead of action sequences, but the application needs to segment these action primitives from the action sequences. UTD-MHAD2 \cite{chen2015utd} is experimented in this case. For small-scale datasets, we performed data augmentation on training data. Besides, PKU-MMD \cite{liu2017pku}, a large public dataset, is also used in our experiment to enrich our data types and increase the credibility of our experiments.

\begin{table*}[ht]
  \centering
  \renewcommand\arraystretch{1.5}
  \caption{Distribution of training data and test data for the three datasets.}
  \label{tab:my-table}
  \begin{tabular}{cccccc}
  \hline
  \multirow{2}{*}{Dataset}  & Data & \multicolumn{2}{c}{\#Training} & \multicolumn{2}{c}{\#Test} \\
  & Augmentation & Subjects & Samples & Subjects & Samples \\ \hline
  \multirow{2}{*}{Tai Chi}   & $\times$  & 1                    & 2                & 2                   & 18              \\
  &  \checkmark & 1                    & 1024                & 2                & 18              \\ \hline
  \multirow{2}{*}{UTD-MHAD2} & $\times$ & 2                    & 2                & 4                   & 28 \\
  & \checkmark & 2                    & 1024                & 4                & 28              \\ \hline
  PKU-MMD                    & $\times$ & 10                   & 775                & 3                 & 234 \\ \hline     
  \end{tabular}
\end{table*}

In this paper, a cross-subject setting is employed in all experiments. Table 1 shows the distribution of training and test data in subsequent experiments on three datasets.

\subsubsection{Tai Chi dataset}
Tai Chi dataset is a self-established dataset for the S-TAS task. Tai Chi actions come from Wu Style Tai Chi, which is one of the most popular genres of Tai Chi. The dataset contains twenty action sequences performed by a male subject and a female subject. Each action sequence has ten action primitives, and each subject repeats the action sequence ten times. Before each recording, the subject was instructed to perform according to the script, but no specific details were given on how to execute the motions. Skeleton data was collected by a wearable inertial sensor system. The skeleton with 18 joints is used in the experiment, as shown in Fig 5(a).

For data augmentation, we select 2 action sequences performed by \textit{subject 1} as training data. Each Tai Chi action sequence has 10 action primitives, so the number of action categories is 10. Due to the coherence of Tai Chi, we retain the order of Tai Chi motions when performing data augmentation. Therefore, according to Equation 2, the number of training samples increases to 1024 (\(2^{10}\)) after data augmentation. The remaining 18 action sequences are used as test data.

\subsubsection{UTD-MHAD2}
UTD-MHAD2 includes ten actions collected by Kinect v2. Three female and three male subjects were asked to perform these ten actions. Each subject repeated actions five times. For the skeleton data, 25 joints were recorded. The skeleton shows in Fig 5(b).

Action categories 1 to 10 of each repetition of each subject are combined into an action sequence with the order. Then set the synthesized frame to category 0, so the number of action categories is 11. In this way, we form a total of 30 (\(5\times 6\)) action sequences. Without data augmentation, we select an action sequence from \textit{subject 1 and 4} respectively as training data, and the remaining 28 action sequences are used as test data. Similar to Tai Chi dataset, after data augmentation, the number of training samples increases from 2 to 1024.

\subsubsection{PKU-MMD}
PKU-MMD is a large-scale benchmark. It covers a wide range of complex human activities with annotated information. We use phase 2 in the dataset. Phase 2 contains 2000 short videos in 49 action categories, performed by 13 subjects in three camera views. Each video contains approximately 7 action instances. The skeleton consists of 25 joints, which are evaluated from the RGB videos and depth maps, also shown in Fig 5(b).

Following the fixed train/test partition for cross-subject experiments of PKU-MMD, we set 775 samples of 10 subjects (\textit{subject 1, 4, 5, 6, 8, 9, 10, 11, 12, 13}) as training data and 234 samples of other 3 subjects as test data.

\subsection{Metrics}
Similar to \cite{farha2019ms, chen2020action, filtjens2022skeleton}, we report the frame-wise accuracy (\(Acc\)) and segment-wise F1 score with the thresholds 0.1, 0.25, 0.5 (F1@{10,25,50}) as evaluation metrics. 

Suppose we have an action sequence included \(H\) frames, and \(h\) frames are classified correctly, then the frame-wise accuracy \(Acc\) is calculated as:
\begin{equation}
Acc = \frac{h}{H}
\end{equation}

Frame-wise accuracy focuses on the classification of each frame. If more frames are labeled correctly, \(Acc\) can get a higher score. But this metric is more sensitive to long action primitives. Therefore, we introduce the F1 score proposed in \cite{lea2017temporal} as a segment-wise metric. F1 score penalizes over-segmentation error and is dependent on the number of action primitives in the action sequences. We calculate whether each predicted action segment is a true positive (\(TP\)) or false positive (\(FP\)) by comparing its IoU with the threshold \(k\).  F1@k can be given by following equations:
\begin{equation}
precision@k = \frac{TP@k}{TP@k + FP@k}
\end{equation}
\begin{equation}
recall@k = \frac{TP@k}{TP@k + FN@k}
\end{equation}
\begin{equation}
F1@k = \frac{2 * (precision@k) * (recall@k)}{(precision@k) +(recall@k)}
\end{equation}

\subsection{Experiment results}
In this paper, we propose two methods to improve the results of the FS-TAS task: synthesizing action sequences to augment the training data and applying CTC to improve the temporal alignment of the predicted results and ground truth. Therefore, we illustrate ablation experiment results to examine the effects of these two parts separately.

For two small-scale datasets, we do four experiments with different settings. PKU-MMD as a large-scale benchmark does not need data augmentation, so we design two experiments to test the effect of CTC. It is worth mentioning that data augmentation is only performed on the training data but not on the test data. Synthesized training data are used when training the deep models. Section 5.1 presents the details of data augmentation on the dataset, and the assignment of training data and test data in the experiments are recorded in Table 1. If not otherwise specific, the loss function mix ratio \(\lambda\) and \(\mu\) are set to 0.15 and 0.0005 respectively.

\begin{table*}[htbp]
  \centering
  \renewcommand\arraystretch{1.5}
  \caption{Performance of Data Augmentation and CTC on the three datasets.}
  \begin{tabular}{ccccccc}
      \hline
  Dataset & Data Augmentation & CTC & Acc & \multicolumn{3}{c}{F1@\{10,25,50\}} \\ \hline
  \multirow{4}{*}{Tai Chi} & $\times$ & $\times$ & 0.776 & 0.735 & 0.697 & 0.606 \\
                           & $\times$  & \checkmark & 0.782 & 0.726 & 0.689 & 0.611 \\
                           & \checkmark & $\times$ & 0.878 & 0.891 & 0.874 & 0.801 \\ 
                           & \checkmark & \checkmark & \textbf{0.895} & \textbf{0.926} & \textbf{0.914} & \textbf{0.846} \\ \hline
  \multirow{4}{*}{UTD-MHAD2} & $\times$ & $\times$ & 0.484 & 0.603 & 0.542 & 0.421 \\
                           & $\times$  & \checkmark & 0.442 & 0.560 & 0.518 & 0.424 \\
                           & \checkmark & $\times$ & 0.866 & 0.762 & 0.746 & 0.705 \\ 
                           & \checkmark & \checkmark & \textbf{0.893} & \textbf{0.837} & \textbf{0.831} & \textbf{0.786} \\ \hline
  \multirow{2}{*}{PKU-MMD} & $\times$ & $\times$ & 0.685 & 0.680 & 0.652 & 0.513 \\
                           & $\times$ & \checkmark & \textbf{0.692} & \textbf{0.699} & \textbf{0.664} & \textbf{0.538} \\ \hline
  \end{tabular}
\end{table*}

Table 2 shows the results of ablation experiments on three datasets, and the best segmentation results are obtained by applying data augmentation and CTC. Our results find clear support that data augmentation can significantly improve action segmentation results on both frame-wise and segment-wise metrics, especially on the UTD-MHAD2. By expanding the training data, the overfitting of deep models can be effectively avoided. Due to data augmentation, we got excellent results using 10\% of the training data on the Tai Chi dataset and 6.7\% of the training data on the UTD-MHAD2. Another finding was that adding CTC in the network architecture can provide a better temporal alignment, which is suggested on the segment-wise metric F1@\{10,25,50\}. It leads to good results, even if the improvement is negligible compared to data augmentation. It is interesting to note that CTC performs better with conducting data augmentation. We speculate that this might be due to the model's need for large amounts of training data, and the performance of CTC depends on the optimal fit of the deep model. Overall, our methods can effectively improve the action segmentation results on the FS-TAS task.

\subsubsection{CTC study}
In this section, we discuss the impact of adding CTC in the network architecture by analyzing the value of parameter \(\mu\) and conducting an ablation experiment of loss functions. After that, we compare the results of our network with state-of-art methods. For simplicity, we use the dataset after data augmentation and only observe the changes caused by CTC.

\textbf{Impact of \(\mu\).} The effect of CTC loss on the segmentation results is controlled by parameter \(\mu\). Therefore, we select the different value of \(\mu\) to observe how it affects the performance of CTC. We experiment \(\mu = 0.0001, 0.0005, 0.001, 0.01\) on the three datasets.

\begin{table*}[ht]
  \centering
  \renewcommand\arraystretch{1.5}
  \caption{Impact of \(\mu\) on the three datasets.}
  \begin{tabular}{ccccccc}
  \hline
  \multirow{2}{*}{\(\mu\) value} & \multicolumn{2}{c}{Tai Chi} & \multicolumn{2}{c}{UTD-MHAD2} & \multicolumn{2}{c}{PKU-MMD} \\
                 & Acc   & F1@50 & Acc   & F1@50 & Acc   & F1@50 \\ \hline
  \(\mu=0.0001\) & 0.873 & 0.792 & 0.883 & \textbf{0.800} & 0.687 & 0.523 \\ 
  \textbf{\(\mu=0.0005\)} & \textbf{0.895} & \textbf{0.846} & \textbf{0.893} & 0.786 & \textbf{0.692} & 0.538 \\
  \(\mu=0.001\) & 0.892 & 0.785 & 0.884 & 0.761 & 0.686 & \textbf{0.544} \\ \hline
  \end{tabular}
  \label{tab:mu}
\end{table*}

As shown in Table \ref{tab:mu}, optimal F1@50 metrics correspond to different optimal \(\mu\) values, this may be due to the different intensity of alignment requirements. For example, action sequences in UTD-MHAD2 include motions arranged in order, so they have low requirements for temporal alignment and are more suitable for a small \(\mu\) value. In contrast, action sequences in the PKU-MMD contain actions that are not fixed in category and order, appropriate for a big \(\mu\) value. Fortunately, the overall impact is small when \(\mu\) fluctuates in a small range (0.0001, 0.001). Therefore, considering everything, we set \(\mu=0.0005\) in all experiments.

\textbf{Ablation experiment of loss function.} To identify the relationship between three loss functions, we conduct an ablation experiment of loss functions on the three datasets.

\begin{table*}[ht]
  \renewcommand\arraystretch{1.5}
  \centering
  \caption{Ablation experiment of loss functions on the three datasets.}
  \begin{tabular}{ccccccccc}
  \hline
  \multicolumn{3}{c}{Loss Function} & \multicolumn{2}{c}{Tai Chi} & \multicolumn{2}{c}{UTD-MHAD2} & \multicolumn{2}{c}{PKU-MMD} \\ 
  \(\mathcal{L}_{cls}\)& \(\mathcal{L}_{T-MSE}\)& \(\mathcal{L}_{CTC}\)& Acc & F1@50 & Acc & F1@50 & Acc & F1@50\\ \hline
  \checkmark &  &  & 0.893 & 0.846 & 0.614 & 0.461 & 0.683 & 0.512 \\
  \checkmark & \checkmark &  & 0.878 & 0.801 & 0.866 & 0.705 & 0.685 & 0.513 \\
  \checkmark & \checkmark & \checkmark & \textbf{0.895} & \textbf{0.846} & \textbf{0.893} & \textbf{0.786} & \textbf{0.692} & \textbf{0.538} \\ \hline
  \end{tabular}
\end{table*}

As can be seen from Table 4, all three datasets get the best experimental results when all loss functions are conducted. Extensive results were carried out to find support for the necessity of three loss functions. Also, from the results, it is clear that adding CTC loss improves the temporal alignment, which ties well with previous studies in wherein speech recognition field. Another finding is that each dataset has different sensitivities to the three loss functions. For example, only using the CE loss function can achieve similar results on the Tai Chi dataset as applying all loss functions. A possible explanation for this might be because of the characteristics of actions. Tai Chi action has strong coherence and is not prone to over-segmentation. Taken together, our experiments indicate that CTC loss can improve the TAS results.

\textbf{Compare with State-of-the-art methods.} Table 5 compares the performance of our model (Figure 4(d)) with state-of-the-art methods and baselines, including ST-GCN (Figure 4(a)), MS-TCN (Figure 4(b)), MS-GCN (Figure 4(c)). For a fair comparison, we evaluated these methods on the PKU-MMD, which is widely used in TAS research.

\begin{table}[ht]
  \centering
  \renewcommand\arraystretch{1.5}
  \caption{Comparison on the PKU-MMD.}
  \begin{tabular}{ccc}
  \hline
  Method     &  Acc  & F1@50 \\ \hline
  SVM        & 0.452 & 0.067 \\
  Bi-LSTM    & 0.565 & 0.254 \\
  Online-LSTM\cite{carrara2019lstm}& 0.577 & 0.233 \\
  ST-GCN\cite{yan2018spatial}     & 0.649 & 0.155 \\
  MS-TCN\cite{farha2019ms}     & 0.655 & 0.463 \\
  MS-GCN\cite{filtjens2022skeleton}     & 0.685 & 0.516 \\ \hline
  Ours       & \textbf{0.692} & \textbf{0.538} \\ \hline
  \end{tabular}
\end{table}

Methods for comparison can be roughly divided into three categories: (1) To compare with the traditional methods, Support Vector Machines (SVM) is chosen as a representative. SVM is more fitting for action recognition, so we cut the action sequences every ten frames as the input of these models. (2) Methods based on RNN or LSTM, which are commonly used for sequence learning. For the Bi-LSTM model, we set the number of layers as 3. Each layer has 128 hidden dimensions. (3) baseline models for our method.

It can be seen in Table 5 that our model achieves state-of-art results on the PKU-MMD v2. LSTM-based models pay more attention to temporal features, emphasizing contextual relationships. In contrast, ST-GCN focuses on extracting spatial features from the skeletons. Therefore, ST-GCN performs better on the frame-wise metric, and the LSTM-based models act better on the segment-wise metric. Further, both MS-TCN and MS-GCN are designed for action segmentation tasks, but MS-TCN is more suitable for video-based data, and MS-GCN is better for skeleton-based data. Overall, slightly superior results are achieved with our method.

\subsubsection{Data augmentation study}
In this section, we design experiments to measure the function of the number of samples, number of subjects, and diversity of action sequences on the results when performing data augmentation.

\textbf{Impact of the number of samples.} First, we investigate the relationship between segmentation results and the number of samples. The Tai Chi dataset is used to conduct this experiment because each subject repeats it enough times so that we can increase training samples but not add subjects. We separately take 2, 4, 6, and 8 action sequences performed by \textit{Subject 1} as training data. To ensure the fairness of the comparison, no matter how many samples were used, we augment 1024 action sequences as training data.

\begin{table*}[ht]
  \centering
  \renewcommand\arraystretch{1.5}
  \caption{Effect of the number of samples on the Tai Chi dataset.}
  \begin{threeparttable}
  \begin{tabular}{cccccc}
      \hline
  Number of samples & Data Augmentation & Acc & \multicolumn{3}{c}{F1@\{10,25,50\}} \\ \hline
  \multirow{2}{*}{2 (s1)\tnote{1}} & $\times$ & 0.782 & 0.726 & 0.689 & 0.611 \\
                            & \checkmark & 0.895 & 0.926 & 0.914 & 0.846 \\ \hline
  \multirow{2}{*}{4 (s1)} & $\times$ & 0.877 & 0.915 & 0.895 & 0.817 \\
                            & \checkmark & 0.922 & 0.976 & 0.963 & 0.895 \\ \hline
  \multirow{2}{*}{6 (s1)} & $\times$ & 0.860 & 0.893 & 0.855 & 0.802 \\
                            & \checkmark & 0.915 & 0.936 & 0.929 & 0.831 \\ \hline
  \multirow{2}{*}{8 (s1)} & $\times$ & 0.896 & \textbf{0.964} & \textbf{0.946} & 0.875 \\
                            & \checkmark & \textbf{0.934} & 0.923 & 0.923 & \textbf{0.923} \\ \hline
  \end{tabular}
  \begin{tablenotes}
  \footnotesize
  \item[1]{Take two samples performed by \textit{Subject 1} as training data. S stands for \textit{Subject}.}
  \end{tablenotes}
  \end{threeparttable}
\end{table*}

As Table 6 shows, metrics show an upward trend with the increase in training samples. The smaller the number of samples, the more we can benefit from data augmentation. However, the observed difference between four samples and eight samples with data augmentation in this study is not significant, so blindly increasing the number of samples may have only a few gains. In contrast, we should focus more on the quality of the training samples rather than the quantity. This may also explain why the result with six samples is worse than four samples. Sample quality is more important than the number of samples.

\textbf{Impact of the number of subjects.} Second, we measure the performance of data augmentation with the different number of subjects. In this study, UTD-MHAD2 is used because it contains enough subjects for experiments. We evaluate 2, 3, and 4 subjects as training data, \textit{Subject 1 and 4} for two subjects, select \textit{Subject 1, 4, and 5} for three subjects, select \textit{Subject 1, 2, 4, and 5} for four subjects. Same as the previous experimental setup, 1024 synthetic action sequences are used as training data.

\begin{table*}[ht]
  \centering
  \renewcommand\arraystretch{1.5}
  \caption{Comparing different number of subjects on the UTD-MHAD2.}
  \begin{threeparttable}
  \begin{tabular}{cccccc}
  \hline
  Number of Subjects & Data Augmentation & Acc & \multicolumn{3}{c}{F1@\{10,25,50\}} \\ \hline
  \multirow{2}{*}{2 (s1,s4)\tnote{1}}     & $\times$ & 0.442 & 0.560 & 0.518 & 0.424 \\
                         & \checkmark & 0.893 & 0.837 & 0.831 & 0.786 \\ \hline
  \multirow{2}{*}{3 (s1,s4,s5)}     & $\times$ & 0.523 & 0.573 & 0.506 & 0.411 \\
                         & \checkmark & 0.914 & 0.882 & 0.875 & 0.865 \\ \hline
  \multirow{2}{*}{4 (s1,s2,s4,s5)}     & $\times$ & 0.739 & 0.818 & 0.751 & 0.628 \\
                         & \checkmark & \textbf{0.933} & \textbf{0.921} & \textbf{0.907} & \textbf{0.893} \\ \hline
  \end{tabular}
  \begin{tablenotes}
      \footnotesize
      \item[1]{Take one sample performed by \textit{Subject 1 and 4} as training data. S stands for \textit{Subject}.}
      \end{tablenotes}
      \end{threeparttable}
\end{table*}

Table 7 provides the experimental results with the different number of subjects. Obviously, there is a significant positive correlation between results and the number of subjects. By increasing the number of subjects, more diverse training data was available, which can help the model learn more discriminative features. Therefore, we can get better segmentation results by adding subjects to the training data and still benefit from data augmentation in this case.

\textbf{Effect of synthetic action sequence diversity.} Third, we evaluate the impact of action order on segmentation when synthesizing action sequences. An unordered synthetic action sequence is formed by combining various action primitives without considering the order of actions. Considering the strong continuity of the Tai Chi action sequence, it is not suitable for disrupting the order of motions. Therefore, we use UTD-MHAD2 for this study. For ease of comparison, the number of training samples is expanded to 1024 through data augmentation, regardless of with or without order.

\begin{table*}[ht]
  \centering
  \renewcommand\arraystretch{1.5}
  \caption{Performance of DA on ordered or unordered synthetic action sequence.}
  \begin{tabular}{cccccc}
  \hline
  Train data with order & Test data with order & Acc   & \multicolumn{3}{c}{F1@\{10,25,50\}} \\ \hline 
  \checkmark & $\times$ & 0.204 & 0.140 & 0.125 & 0.100 \\ 
  $\times$   & $\times$ & \textbf{0.514} & \textbf{0.457} & \textbf{0.447} & \textbf{0.425} \\ \hline
  $\times$   & \checkmark & 0.475 & 0.441 & 0.424 & 0.414 \\ 
  \checkmark & \checkmark & \textbf{0.893} & \textbf{0.837} & \textbf{0.831} & \textbf{0.786} \\ \hline
  \end{tabular}
\end{table*}

Table 8 reports the experimental results under different conditions. As we expected, the best results are obtained when the training and test data have the same form. It means that we have to ensure that the synthesized data form is consistent with real data when performing data augmentation. Notably, the result is significantly worse when test data is out of order. A possible explanation for this result may be the insufficient diversity of training data. We only use two samples, 2/30 of the total data, for data augmentation. It may be sufficient for ordered test data since the order of actions is a feature to help the model learn, but not enough for unordered test data. In this case, segmentation results can be improved by adding new samples or subjects, as introduced previous in this section.

\section{Conclusion and Future Work}
\label{sec6}
In this paper, we present a data augmentation method to address the insufficient number of training samples in the few-shot skeleton-based action segmentation task. At the same time, we propose adding CTC to a baseline architecture for skeletal action segmentation. Extensive experimental results show that both our proposed methods are effective. A limitation of this study is that when the gap between skeletons in the two frames is too large, it is difficult to achieve a perfect synthesis due to the limitations of action interpolation. Therefore, in future work, we will explore better motion interpolation methods, so that data augmentation can not only augment the amount of data but also help improve the diversity of data.

\section*{Acknowledgments}
This work is supported by the National Natural Science Foundation of China (No. 61876054). Tai Chi dataset is available at \url{https://hit605.org/projects/taichi-data/}

\bibliographystyle{unsrt}  
\bibliography{paper.bib}

\end{document}